\documentclass[letterpaper]{article} 
\usepackage{aaai2026}  
\usepackage{times}  
\usepackage{helvet}  
\usepackage{courier}  
\usepackage[hyphens]{url}  
\usepackage{graphicx} 
\usepackage{natbib}  
\usepackage{caption} 
\usepackage{algorithm}
\usepackage{algorithmic}
\usepackage{amsmath}

\usepackage{newfloat}
\usepackage{listings}
\DeclareCaptionStyle{ruled}{labelfont=normalfont,labelsep=colon,strut=off} 
\floatstyle{ruled}
\newfloat{listing}{tb}{lst}{}
\floatname{listing}{Listing}
\title{	
A Robust State Filter Against Unmodeled Process And Measurement Noise}
\author {
Weitao LIU
}
\affiliations {
}

\begin{document}

\maketitle

\begin{abstract}
This paper introduces a novel Kalman filter framework designed to achieve robust state estimation under both process and measurement noise. Inspired by the Weighted Observation Likelihood Filter (WoLF), which provides robustness against measurement outliers, we applied generalized Bayesian approach to build a framework considering both process and measurement noise outliers. 
\end{abstract}

\section{Introduction}

The Kalman Filter (KF) is a cornerstone of estimation theory and stochastic filtering. For linear dynamical systems with Gaussian process and measurement noise, the KF provides the optimal minimum-mean-square-error (MMSE) state estimate via a recursive predict–update cycle. \citep{Kalman1960} However, KF has several assumptions that may lead to performance degradation when violated:

\paragraph{Linear assumption of state transition model and measurement model} A very common approach is to use Extended Kalman Filter (EKF), which uses first-order approximation to avoid nonlinearity. Another common class of practice is to use sigma-point methods to perform deterministic sampling, with Unscented Kalman Filter (UKF) and Cubature Kalman Filter (CKF) as two widely used novel variants. \citep{Van2004}

\paragraph{Assumption of Gaussian measurement noise} KF and EKF assume Gaussian noise and are derived from a least-squares (MMSE) criterion, so it can be sensitive to measurement outliers. \citep{8398426, duran2024} There's a large quantity of literature aiming at addressing this measurement model misspecification problem. One common approach is to use different measurement noise model. This includes variants using Student's t, Laplace, or other mixture models. \citep{8009803, 7812899, gong2023, s24092720,8214971,7527863}. Another common approach is to use M-estimation. This includes variants using Huber's loss to blend l1 and l2 norms. \citep{Karl2007,4047553, 5371933}. Alternatively, Kalman Smoothing methods are also proposed to use a post-processing to refine past estimations, therefore reducing impact of non-Gaussian measurements. The variants include simple ones like fixed-lag and fixed-point smoother \citep{6243075}, Rauch–Tung–Striebel (RTS) smoothers \citep{5371933} and other Robust smoothers \citep{aravkin2013}

\paragraph{Correct pre-determined covariance matrices} In KF setting, the process noise matrix Q and measurement noise matrix R is pre-determined, which often deviates from true ones in common practice. When the covariance of noises are not correct or the noises are time-varying signals, the performance of standard KF or EKF may be poor or even diverge. Multiple methods have been proposed to estimate covariance \citep[e.g.][]{8273755,8996637,Zhang_Wang_Sun_Gao_2017,rs15174125,chen2015}, and handling time-varying noise \citep[e.g.][]{JIA2020368}. .

There are also other simulation-based methods like Sequential Monte-Carlo (SMC) methods, also known as Particle Filters (PF), which also falls into the general category of Bayesian Filtering. SMC methods are much more flexible but requires more computational power. \citep{Doucet2001} As an extension of PF and KF, Ensemble Kalman Filter (EnKF), which represents the state distribution using an ensemble of simulations, is proposed as a novel framework being adaptive like Monte-Carlo methods, but reduces computational complexity compared to SMC methods as it keeps similar assumptions as KF. \citep{Evensen_2003}

While many Kalman filter variants address one of these problems, achieving robustness to both innovations and outliers is challenging due to trade-offs between responding speed and robustness against outliers. Inspired by the WoLF approach by \citet{duran2024}, we propose a Kalman Filter Framework that adaptively weights process noise and measurement noise, effectively addressing both innovative and additive outliers.

Our work offers several advantages: 1. Easy Implementation and Efficiency: The method has a structure similar to the standard Kalman Filter, ensuring straightforward implementation and low computational cost. 2. Robustness to Misspecification: It adapts effectively to inaccuracies in specifying process and measurement noise. 3. Extensibility: The approach can be readily extended to other Kalman Filter variants.

\section{Problem formulation}

In this section, we formulate the state-space model setting in KF setting. In state-space filtering problems, we are interested in a state vector \(x\in R^{m}\) while only observing a corresponding observation vector \(y\in R^{d}\). Given an initial state \(x_0\), the state-space model is defined as:
\begin{align}
    x_t &= f_t(x_{t-1},u_t)\\
    y_t &= h_t(x_t,v_t)
\end{align}
Here, \(x_t,y_t\) are the state vector and the measurement vector for each time. \(f_t\) is the function that transforms \(x_{t-1}\) to \(x_t\), with the effect from a process noise vector \(u_t\sim\mathcal{N}(0,Q_t)\). \(h_t\) is the function that maps the state vector \(x_t\) to the corresponding measurement vector \(y_t\), with the effect from a measurement noise vector \(v_t\sim\mathcal{N}(0,R_t)\).

This also yields the standard Bayesian prediction:
\begin{align}
    &p(x_t|y_{1\cdots t-1})\\ = &\iint p(x_t|x_{t-1},u_t)p(x_{t-1}|y_{1\cdots t-1})p(u_t)\; \mathrm{d} u_t \mathrm{d} x_{x-1}
\end{align}
and update:
\begin{align}
    p\left(x_t | y_{1: t}\right)=\frac{p\left(y_t | x_t\right) p\left(x_t | y_{1: t-1}\right)}{p\left(y_t | y_{1: t-1}\right)}
\end{align}
where
\begin{align}
    p\left(y_t | y_{1: t-1}\right)=\iint p\left(y_t | x_t,v_t\right) p\left(x_t | y_{1: t-1}\right)\;\mathrm{d}v_t\mathrm{d} x_t
\end{align}
\subsection{Kalman filter}
Kalman filter assumes additive zero-mean Gaussion Noise and linear \(f_t,h_t\). Then the model is defined as:
\begin{align}
    x_t &= F_t x_{t-1}+u_t\\
    y_t &= H_t x_{t-1}+v_t
\end{align}
Here \(x_t,y_t\) represents the random variables, and \(u_t\sim\mathcal{N}(0,Q_t),v_t\sim\mathcal{N}(0,R_t)\). Then derived from the standard Bayesian formula, the prediction step is defined as:
\begin{align}
    x_{t|t-1} &= F_t x_{t-1|t-1}\\
    \Sigma_{t|t-1} &= F_t \Sigma_{t-1|t-1}F_t^T+Q_t
\end{align}
and the update step
\begin{align}
    \Sigma_t^{-1} & =\Sigma_{t \mid t-1}^{-1}+H_t^{\top} R_t^{-1} H_t \\
    K_t & =\Sigma_t H_t^{\top} R_t^{-1} \\
    x_{t|t} & =x_{t \mid t-1}+K_t\left(y_t-H_tx_t\right)
\end{align}
This is also known as the information form of Kalman Filter. When addressing nonlinearity, EKF uses first-order approximation of \(F_t,H_t\), sigma-points variants like UKF or CKF propogate sigma points to get estimatation.
\subsection{Weighted observation likelihood filter}
\citet{duran2024} proposed a robust KF framework called  Weighted observation likelihood filter (WoLF) inspired by Generalized Bayes. In the update step:
\begin{align}
    p\left(x_t | y_{1: t}\right)\propto p\left(y_t | x_t\right) p\left(x_t | y_{1: t-1}\right)
\end{align}
the posterior is changed to a generalized posterior:
\begin{align}
    q\left(x_t | y_{1: t}\right)\propto \exp\left(-\ell_t(y_t)\right) q\left(x_t | y_{1: t-1}\right)
\end{align}
with
\begin{align}
    \ell_t(y_t)=-W^2(y_t,\hat{y_t})\log q(y_t|x_t)
\end{align}
with $W:\mathbf{R}^2\to\mathbf{R}$ a weighting function that controls the impact of new measurement. 

When using linear approximation and accepting Gaussian assumptions, the modified update step can be formulated as follow:
\begin{align}
    \hat{y_t} &= h_t(x_{t|t-1})\\
    \omega_t &= W(y_t,\hat{y_t})\\
    \Sigma_t^{-1} & =\Sigma_{t \mid t-1}^{-1}+\omega_t^2H_t^{\top} R_t^{-1} H_t \\
    K_t & =\omega_t^2\Sigma_t H_t^{\top} R_t^{-1} \\
    x_{t|t} & =x_{t \mid t-1}+K_t\left(y_t-\hat{y_t}\right)
\end{align}

By setting outlying measurements lower weight, the filter gains robustment against measurement outliers by reducing information gain. However, when unmodeled state change occurs, it also limits the filter from converging fast to the new state.
\subsection{Robust Loss-based Filter}
To address this issue, we propose a mathematically guaranteed framework that considers both process and measurement noise. Relevent proof can be found in Appendix, here we provide the intuitive formulation.

Recall the standard Bayesian prediction:
\begin{align}
    p\left(x_t | y_{1: t}\right)\propto p\left(y_t | x_t\right) p\left(x_t | y_{1: t-1}\right)
\end{align}
here we set both part related to $Q,R$ to generalized posterior:
\begin{align}
    q\left(x_t | y_{t}\right)\propto \exp\left(-\ell_t(y_t)\right)\exp\left(-\gamma_t(x_t,Q)\right)
\end{align}
When using linear approximation and accepting Gaussian assumptions, we can express the prediction step as below:
\begin{align}
    x_{t|t-1} &= F_t x_{t-1|t-1}\\
    \Sigma_{t|t-1} &= \Lambda_t F_t \Sigma_{t-1|t-1}F_t^T\Lambda_t^T+\Omega_t Q_t \Omega_t^T
\end{align}
and same update step as in WoLF:
\begin{align}
    \hat{y_t} &= h_t(x_{t|t-1})\\
    \omega_t &= W(y_t,\hat{y_t})\\
    \Sigma_t^{-1} & =\Sigma_{t \mid t-1}^{-1}+\omega_t^2H_t^{\top} R_t^{-1} H_t \\
    K_t & =\omega_t^2\Sigma_t H_t^{\top} R_t^{-1} \\
    x_{t|t} & =x_{t \mid t-1}+K_t\left(y_t-\hat{y_t}\right)
\end{align}
Our Robust Loss-based Filter (RoLF) enhances the Kalman Filter by adding two user-defined matrices, \(\Lambda_t\) and \(\Omega_t\), to improve robustness. \(\Lambda_t\) scales the propagated covariance \(\Sigma_{t-1|t-1}\) through the transition matrix \(F_t\), while \(\Omega_t\) adjusts the process noise \(Q_t\). Both are derived from user-specified loss functions, allowing RoLF to handle process outliers and adapt to noise misspecifications effectively. Much previous work aimed at achieving robustness to process noise in Kalman filtering can be viewed as special cases within our RoLF framework.

\section{Experiment}
In this section, we provide a simple scenario of estimating the position of a object moving on 2D plane with GARCH process noise in velocity.
\subsection{Loss function selection}
In this section, we employ the Mahalanobis-based weighting function for the measurement loss:
\begin{align}
    W(y_t, \hat{y}_t) = \left( 1 + \frac{\| R_t^{-1/2} (y_t - \hat{y}_t) \|_2^2}{c^2} \right)^{-1/2}
\end{align}
Previous works have used Mahalanobis-distance–based weights and p-Huber / Mahalanobis three-stage weight functions for robust filtering, showing improved empirical performance compared with other kernels\citep{gao2019robust_ckf,hu2022p_huber,javanfar2023_measurement_outlier}.
For process-noise robustness we adapt ideas from the Strong Tracking Filter (STF) literature\citep{liu2020_strong_tracking_ckf,zhou1996_strong_tracking}:
\begin{align}
    V_t&=\rho H_t\Sigma_{t-1|t-1}H_t^T+R^t+(1-\rho)y_t y_t^T\\
    \theta &= \max(1,\frac{tr(V_t-R_t-H_tQ_{t-1}H_t^T)}{tr(H_tF_{t-1}\Sigma_{t-1|t-1}F_{t-1}^TH_t^T)})\\
    \Sigma_{t|t-1} &= \theta\Lambda_t F_t \Sigma_{t-1|t-1}F_t^T\Lambda_t^T+\Omega_t Q_t
\end{align}
where $\rho$ is a user-specified smoothing parameter, and $y_t$ is the difference between predicted and real measurement.

\subsection{Problem Setting}
The dataset represents a 2D Generalized Autoregressive Conditional Heteroskedasticity (GARCH) process with mixture measurements, simulating a dynamic system with unknown large perturbance in process. The state vector includes position and velocity components $(x, vx, y, vy)$, and perturbed by Gaussian noise with fixed variances for position and GARCH-driven variances for velocity. The measurement noise consists primarily of Gaussian noise, with a small probability of large-scale outliers.

\includegraphics[width=0.9\columnwidth]{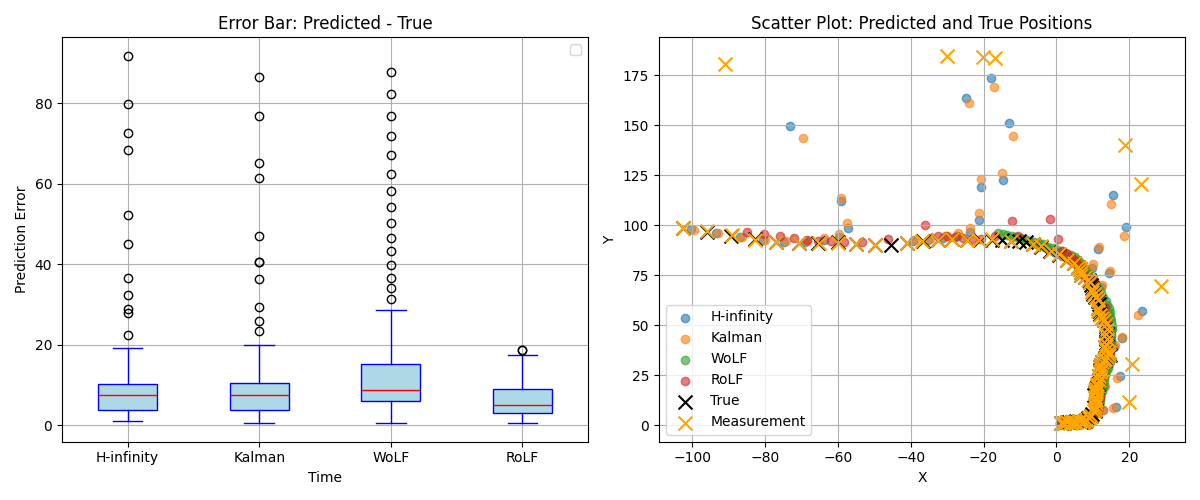}
\label{fig:plot1}

To highlight the performace in handling extreme outliers caused by outliers, we also evaluating the 5\% largest losses highlights the filter's robustness, which is critical for assessing performance in worst-case scenarios.
\includegraphics[width=0.9\columnwidth]{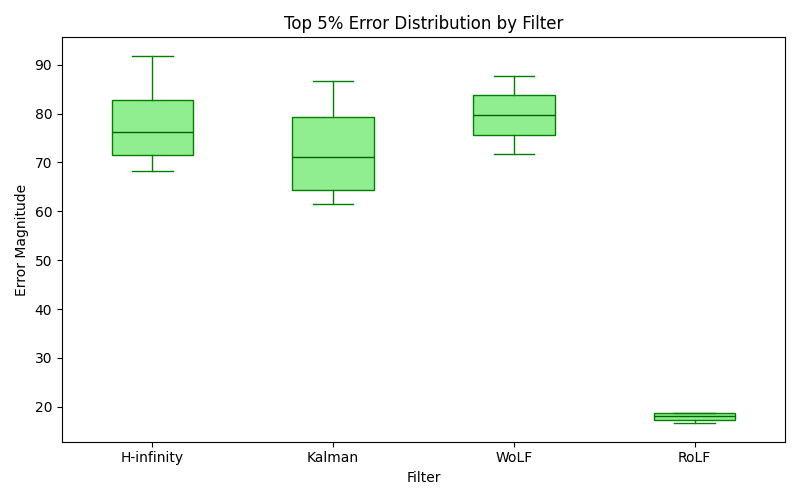}
\label{fig:plot2}

\section{Conclusion}
In this paper, we introduced the Robust Loss-based Filter (RoLF), a novel Kalman filter framework that extends the Weighted Observation Likelihood Filter (WoLF) by incorporating generalized Bayesian principles to address both process and measurement noise outliers. Through our experiments on a 2D GARCH process with mixture measurements, RoLF demonstrated superior robustness, effectively mitigating the impact of outliers in both process and measurement noise while achieving lower worst-case losses compared to standard Kalman filters and WoLF. Evaluating the 5\% largest losses is particularly insightful as it emphasizes the filter's performance in extreme scenarios, where robustness against severe outliers is most critical. Our approach not only maintains computational efficiency and ease of implementation but also offers extensibility to other Kalman filter variants.

\bibliography{aaai2026}

@article{Kalman1960,
    Author = {Kalman, Rudolph Emil},
    Title = {A New Approach to Linear Filtering and Prediction Problems},
    Journal = {Transactions of the ASME--Journal of Basic Engineering},
    Volume = {82},
    Number = {Series D},
    Pages = {35--45},
    Year = {1960}
}

@phdthesis{Van2004,
author = {Van Der Merwe, Rudolph and Wan, Eric A.},
title = {Sigma-point kalman filters for probabilistic inference in dynamic state-space models},
year = {2004},
publisher = {Oregon Health \& Science University},
note = {AAI3129163}
}

@article{Doucet2001,
author = {Doucet, Arnaud and Freitas, Nando and Gordon, Neil},
year = {2001},
month = {01},
pages = {},
title = {An Introduction to Sequential Monte Carlo Methods},
isbn = {978-1-4419-2887-0},
journal = {Sequential Monte Carlo Methods in Practice. Springer, Berlin},
doi = {10.1007/978-1-4757-3437-9_1}
}

@ARTICLE{8398426,
  author={Wang, Hongwei and Li, Hongbin and Fang, Jun and Wang, Heping},
  journal={IEEE Signal Processing Letters}, 
  title={Robust Gaussian Kalman Filter With Outlier Detection}, 
  year={2018},
  volume={25},
  number={8},
  pages={1236-1240},
  doi={10.1109/LSP.2018.2851156}}

@misc{duran2024,
      title={Outlier-robust Kalman Filtering through Generalised Bayes}, 
      author={Gerardo Duran-Martin and Matias Altamirano and Alexander Y. Shestopaloff and Leandro Sánchez-Betancourt and Jeremias Knoblauch and Matt Jones and François-Xavier Briol and Kevin Murphy},
      year={2024},
      eprint={2405.05646},
      archivePrefix={arXiv},
      primaryClass={stat.ML},
      url={https://arxiv.org/abs/2405.05646}, 
}

@INPROCEEDINGS{8009803,
  author={Wang, Hongwei and Li, Hongbin and Zhang, Wei and Wang, Heping},
  booktitle={2017 20th International Conference on Information Fusion (Fusion)}, 
  title={Laplace l1 robust Kalman filter based on majorization minimization}, 
  year={2017},
  volume={},
  number={},
  pages={1-5},
  doi={10.23919/ICIF.2017.8009803}}

@ARTICLE{7812899,
  author={Huang, Yulong and Zhang, Yonggang and Li, Ning and Chambers, Jonathon},
  journal={IEEE Transactions on Aerospace and Electronic Systems}, 
  title={Robust student’s t based nonlinear filter and smoother}, 
  year={2016},
  volume={52},
  number={5},
  pages={2586-2596},
  doi={10.1109/TAES.2016.150722}}

@misc{gong2023,
      title={A Covariance Adaptive Student's t Based Kalman Filter}, 
      author={Benyang Gong and Jiacheng He and Gang Wang and Bei Peng},
      year={2023},
      eprint={2309.09565},
      archivePrefix={arXiv},
      primaryClass={eess.SP},
      url={https://arxiv.org/abs/2309.09565}, 
}

@Article{s24092720,
AUTHOR = {Yu, Benru and Gu, Hong and Su, Weimin},
TITLE = {A Robust Interacting Multi-Model Multi-Bernoulli Mixture Filter for Maneuvering Multitarget Tracking under Glint Noise},
JOURNAL = {Sensors},
VOLUME = {24},
YEAR = {2024},
NUMBER = {9},
ARTICLE-NUMBER = {2720},
URL = {https://www.mdpi.com/1424-8220/24/9/2720},
PubMedID = {38732826},
ISSN = {1424-8220},
DOI = {10.3390/s24092720}
}

@article{Karl2007,
author = {Karlgaard, Christopher D. and Schaub, Hanspeter},
title = {Huber-Based Divided Difference Filtering},
journal = {Journal of Guidance, Control, and Dynamics},
volume = {30},
number = {3},
pages = {885-891},
year = {2007},
doi = {10.2514/1.27968}
}

@INPROCEEDINGS{4047553,
  author={Boncelet, Charles G. and Dickinson, Bradley W.},
  booktitle={The 22nd IEEE Conference on Decision and Control}, 
  title={An approach to robust Kalman filtering}, 
  year={1983},
  volume={},
  number={},
  pages={304-305},
  doi={10.1109/CDC.1983.269847}}

@ARTICLE{8214971,
  author={Huang, Yulong and Zhang, Yonggang and Shi, Peng and Wu, Zhemin and Qian, Junhui and Chambers, Jonathon A.},
  journal={IEEE Transactions on Systems, Man, and Cybernetics: Systems}, 
  title={Robust Kalman Filters Based on Gaussian Scale Mixture Distributions With Application to Target Tracking}, 
  year={2019},
  volume={49},
  number={10},
  pages={2082-2096},
  doi={10.1109/TSMC.2017.2778269}}

@INPROCEEDINGS{7527863,
  author={Huang, Yulong and Zhang, Yonggang and Li, Ning and Mohsen.Naqvi, Syed and Chambers, Jonathon},
  booktitle={2016 19th International Conference on Information Fusion (FUSION)}, 
  title={A robust Student's t based cubature filter}, 
  year={2016},
  volume={},
  number={},
  pages={9-16},
  doi={}}

@ARTICLE{5371933,
  author={Gandhi, Mital A. and Mili, Lamine},
  journal={IEEE Transactions on Signal Processing}, 
  title={Robust Kalman Filter Based on a Generalized Maximum-Likelihood-Type Estimator}, 
  year={2010},
  volume={58},
  number={5},
  pages={2509-2520},
  doi={10.1109/TSP.2009.2039731}}

@INPROCEEDINGS{6243075,
  author={Lu, Xiao and Haixia Wang and Mingchao Li},
  booktitle={2012 24th Chinese Control and Decision Conference (CCDC)}, 
  title={Kalman fixed-interval and fixed-lag smoothing forwireless sensor systems with multiplicative noises}, 
  year={2012},
  volume={},
  number={},
  pages={3023-3026},
  doi={10.1109/CCDC.2012.6243075}}

@misc{aravkin2013,
      title={Sparse/Robust Estimation and Kalman Smoothing with Nonsmooth Log-Concave Densities: Modeling, Computation, and Theory}, 
      author={Aleksandr Y. Aravkin and James V. Burke and Gianluigi Pillonetto},
      year={2013},
      eprint={1301.4566},
      archivePrefix={arXiv},
      primaryClass={stat.ML},
      url={https://arxiv.org/abs/1301.4566}, 
}

@INPROCEEDINGS{8273755,
  author={Akhlaghi, Shahrokh and Zhou, Ning and Huang, Zhenyu},
  booktitle={2017 IEEE Power \& Energy Society General Meeting}, 
  title={Adaptive adjustment of noise covariance in Kalman filter for dynamic state estimation}, 
  year={2017},
  volume={},
  number={},
  pages={1-5},
  doi={10.1109/PESGM.2017.8273755}}

@INPROCEEDINGS{8996637,
  author={Long, Zixuan and Zhang, Xiaoli and Peng, Xiafu and Yang, Gongliu},
  booktitle={2019 Chinese Automation Congress (CAC)}, 
  title={An Improved Adaptive Extended Kalman Filter Used for Target Tracking}, 
  year={2019},
  volume={},
  number={},
  pages={1017-1022},
  doi={10.1109/CAC48633.2019.8996637}}

@article{Zhang_Wang_Sun_Gao_2017, title={Adaptive Cubature Kalman filter based on the variance-covariance components estimation}, volume={15}, DOI={10.1186/s41445-017-0006-z}, number={1}, journal={The Journal of Global Positioning Systems}, author={Zhang, Ya and Wang, Jianguo and Sun, Qian and Gao, Wei}, year={2017}, month={Mar}}

@Article{rs15174125,
AUTHOR = {Yin, Zhihui and Yang, Jichao and Ma, Yue and Wang, Shengli and Chai, Dashuai and Cui, Haonan},
TITLE = {A Robust Adaptive Extended Kalman Filter Based on an Improved Measurement Noise Covariance Matrix for the Monitoring and Isolation of Abnormal Disturbances in GNSS/INS Vehicle Navigation},
JOURNAL = {Remote Sensing},
VOLUME = {15},
YEAR = {2023},
NUMBER = {17},
ARTICLE-NUMBER = {4125},
URL = {https://www.mdpi.com/2072-4292/15/17/4125},
ISSN = {2072-4292},
DOI = {10.3390/rs15174125}
}

@misc{chen2015,
      title={Maximum Correntropy Kalman Filter}, 
      author={Badong Chen and Xi Liu and Haiquan Zhao and José C. Príncipe},
      year={2015},
      eprint={1509.04580},
      archivePrefix={arXiv},
      primaryClass={stat.ML},
      url={https://arxiv.org/abs/1509.04580}, 
}

@article{JIA2020368,
title = {A Novel Robust Kalman Filter With Non-stationary Heavy-tailed Measurement Noise⁎⁎This work was supported in part by the National Natural Science Foundation of China under Grants 61903097 and 61773133, in part by the Fundamental Research Funds for the Central Universities under Grants 3072019CFJ0411 and GK204026025901. Corresponding author is Y. L. Huang.},
journal = {IFAC-PapersOnLine},
volume = {53},
number = {2},
pages = {368-373},
year = {2020},
note = {21st IFAC World Congress},
issn = {2405-8963},
doi = {https://doi.org/10.1016/j.ifacol.2020.12.188},
url = {https://www.sciencedirect.com/science/article/pii/S2405896320304535},
author = {Guangle Jia and Yulong Huang and Mingming B. Bai and Yonggang zhang}
}

@article{Evensen_2003, title={The ensemble Kalman Filter: Theoretical formulation and practical implementation}, volume={53}, DOI={10.1007/s10236-003-0036-9}, number={4}, journal={Ocean Dynamics}, author={Evensen, Geir}, year={2003}, month={Nov}, pages={343–367}}

@article{gao2019robust_ckf,
  title={A Robust Cubature Kalman Filter with Abnormal Observation Identification and Adaptive Scaling},
  author={Gao, B. and others},
  journal={Sensors},
  year={2019},
  volume={19},
  number={23},
  pages={5149},
  doi={}
}

@article{hu2022p_huber,
  title={Robust Estimation in Continuous–Discrete Cubature Kalman Filter},
  author={Hu, H.},
  journal={Applied Sciences},
  year={2022},
  volume={12},
  number={16},
  pages={8167},
  doi={}
}

@article{javanfar2023_measurement_outlier,
  title={Measurement-outlier robust Kalman filter for discrete-time systems},
  author={Javanfar, E.},
  journal={ISA Transactions},
  year={2023},
  doi={}
}

@article{zhou1996_strong_tracking,
  title={Strong tracking Kalman filtering of nonlinear time-varying stochastic systems with coloured noise: application to parameter estimation and empirical robustness analysis},
  author={Zhou, D. H. and Frank, P. M.},
  journal={International Journal of Control},
  year={1996},
  volume={65},
  pages={295--307},
  doi={}
}

@article{liu2020_strong_tracking_ckf,
  title={A Strong Tracking Mixed-Degree Cubature Kalman Filter Method and Its Application in a Quadruped Robot},
  author={Liu, J. and Wang, P. and Zha, F. and others},
  journal={Sensors},
  year={2020},
  volume={20},
  pages={2251},
  doi={}
}

\end{document}